# Is there an aesthetic component of language?

*Harshit Parmar, Jeffrey P. Williams*

Texas Tech University

**Abstract**

Speakers of all human languages make use of grammatical devices to express attributional qualities, feelings, and opinions as well as to provide meta-commentary on topics in discourse. In general, linguists refer to this category as 'expressives'[1] in spite of the fact that defining exactly what 'expressives' are remains elusive. The elusiveness of expressives has given rise to considerable speculation about the nature of expressivity as a linguistic principle. Specifically, several scholars have pointed out the 'special' or 'unusual' nature of expressives vis-à-vis 'normal' or 'natural' morpho-syntax (Diffloth 1972, 1976 & 1980; Dingemanse 2012, 2015 & 2018; Williams 2014; Zwicky and Pullum 1987).

---

[1] Linguistic analyses of ideophones, reduplication, echo words, onomatopoeia, and other non-categorical forms has grown over the past two decades (cf. Voeltz & Kilian-Hatz 2001; McCarthy, Kimper and Mullin 2012; Akita 2015; Iwasaki, Sells & Akita 2017; Ibarretxe-Antuñano 2017; Dingemanse 2012, 2015 & 2018; Haiman 2019; and, Akita and Pardeshi 2019; Williams 2014, 2020, 2021).

It has been speculated that because of their unique traits, expressives are perceived differently in the linguistic stream of understanding and here we conduct a preliminary investigation to validate that using a neuroimaging experiment. It was identified in the results that there are brain regions outside the traditional language processing areas which were involved in interpretation of expressives. In bilinguals, more brain activation was observed when interpreting the expressives in native language as opposed to the secondary language. Although effect size was small, activation regions for expressive perception show an overlap with regions representing creativity and uniqueness indicating that expressives are in fact perceived as a different, rather creative component of the language.

# 1. Introduction

Language processing is a complex task performed by the brain. Studies suggest that a number of functional brain regions are involved in understanding different aspects of language [Hickok, 2009]. Language being an important social aspect of an individual, the language regions have been studied extensively using functional neuroimaging tools. Identification of language regions have been a crucial factor for presurgical procedures as well. For presurgical procedures, language regions have been identified using both task [Benjamin et al., 2017] and resting state [Tie et al., 2014] fMRI paradigms. The temporal and spatial reliability of these language regions has been successfully tested with multiple longitudinal studies [Zhu et al., 2014; Nettekoven et al., 2018].

Within the language regions, aspects of grammar are interpreted in different locations. Semantic processing regions differ from syntactic processing [Rüschemeyer 2005] and so do the nominal and verbal processing regions [Sahin et al., 2006]. But collectively, most of the language processing regions lie either within the Broca's area or the Wernicke's area of the brain. However, some aspects of language are being processed differently by the brain. For example, grammar norm violations are in fact processed partially outside the classic language regions [Hubers et al., 2016]. Along with that, taboo words (cursing or swearing), which are also part of every language, are processed differently from non-taboo words inside the brains of bilinguals [Sulpizio et al., 2019]. Apart from involving more brain regions, processing times for taboo words was also shown to be longer than for neutral words [Sulpizio et al., 2019]. It is thus plausible to think that there could be other aspects of language which are interpreted in a different way inside the brain. One such aspect of language is the category of expressives. Expressives are very prevalent in South Asian languages [cf. Williams 2020] and, most importantly for our study, are thought to belong to a special category of grammar [cf. Diffloth 1972; Zwicky and Pullum 1987].

The proposed study seeks to provide preliminary data to assist in answering the question "Is there an aesthetic component of language?" The idea of an aesthetic component of language was first proposed in modern linguistics by Gérard Diffloth in 1972 in his seminal article "Notes on Expressive Meaning." [Diffloth, 1972]. In that article, Diffloth made the claim that features of grammar, such as expressives, belonged to a discrete and separate category, or component of language that he called 'esthetic.' The esthetic/aesthetic component existed separately from other components of grammar, leading us to speculate that there might be some different localization for the brain activity associated with the production and interpretation of expressives.

In the following section we outline expressivity and provides examples of expressives to give the relevant linguistic background to our neuropsychological study of expressives in Gujarati and Hindi.

*1.1 Expressivity and Expressives*

We can assume, relatively non-controversially, that grammar is comprised of representations guided by principles and rules. The exact shape of these components is still in debate, but overall, principles set the stage for the production of semantic-syntactic strings through an orderly application of a rule component. Expressivity certainly takes part in this sort of structure, also being governed by both principles and rules and shaped by representational constraints.

**1.1.1 Expressivity**

Expressivity is no different from other properties of human language that find articulation in the grammar through principles and rules. Expressivity is governed by a principle of expressivity that is universal in human language, which states the following: a systematic feature of human language is the ability to articulate and communicate perception of natural and social worlds [Williams, forthcoming]. We refer to this feature as the principle of expressivity.

The principle of expressivity is manifested through the grammatical resources of human languages. The grammar of each language has its own set of structures that can be employed by speakers to reflect on perceptions of actions, activities, and social positions of individuals. This accounts for the variation we find across languages in terms of what structurally constitutes expressivity. In this study we examine the principle of expressivity in Gujarati (Indo-Aryan) and Hindi (Indo-Aryan) through a pilot study.

*1.1.1.1 Expressives*

Expressives are shape-shifting forms whose functions cross-cut grammatical categories and classes but, in general, serve to allow the speaker to provide meta-commentary on an argument in the discourse. As the name conveys, expressives allow speakers to 'express' an opinion, an attitude, a perception, or other psychological state regarding a topic in a situated discourse [Williams, 2021]. As Tuvasson states, "[e]xpressives typically package multiple aspects of a sensory event in a single word" [2011: p. 88].

The concept of expressives, sometimes referred to as ideophones, is a relatively young one in linguistics. One of the earliest descriptions of an expressive can be found in the writings of the psychologist-phoneticist Edward Wheeler Scripture (1864-1945). In his *Elements of Experimental Phonetics*, Scripture used the term *ideogram* for the holistic perception of a printed word. He states, '*printed words are perceived in wholes as ideograms and not as combinations (…) … words may be perceived under conditions that exclude any perception of the single* *elements*' [Scripture, 1902: p. 128]. Scripture goes further to create parallel between the 'image' of the printed word and that of the auditory word:

It may be suggested that auditory words and phrases from 'ideophones' just as printed ones form 'ideograms'. The further distinctions may be made of ideograms and ideophones into sensory (visual words and auditory words) and motor ones (written words and spoken words). In all probability the most prominent features of a phonetic unit are first perceived, and the details are gradually filled in [Scripture, 1902: p. 132].

For Scripture then, an expressive is a unit of sound that is perceived as a whole rather than as a combination of some parts; and this whole represents one idea. This condition is what has driven linguists to consider expressives as somehow special in terms of their linguistic categorization.

There are several types or forms of expressives found in the grammars of human languages. It is beyond the scope of the present paper and reported study to go into these in any detail. The particular kind of expressives that we employ in our study are reduplications and echo words and we will define and illustrate those in the following sections.

*1.1.1.2 Reduplication*

Reduplication is a word formation process involving the copying of a base or root form and collocating the two together. Theoretically, there is disagreement as to whether reduplication is a form of compounding or a form of affixation. That debate need not trouble us here and is not relevant for the present study. Examples of reduplication from our data set include the following examples from Gujarati and Hindi respectively:

આ ઘાવ **ધીમે ધીમે** રુજાઈ જશે.

*Aa ghaav **dheeme dheeme** rujai jase.*

The wound will heal slowly.

मुझे **गरम गरम** पकोड़े खाने हैं ।

*Muje **garam garam** pakode khaane hai.*

I want to eat hot fritters.

In each example, the first sentence is a sentence with expressive (**highlighted**) in the native script, the second one is the Romanized pronunciation, and the third sentence is the translation in English.

*1.1.1.3 Echo words*

Echo-words are a special word formation process involving the formation/generation of rhyming words through copying, change, and collocation. Rhyming is a tricky term to use in a universal fashion since the conceptualization of what it means to rhyme is governed by local grammatical conditions. Abbi defines an echo word as *'. . . a partially repeated form of the base word – partially in the sense that either the initial phoneme (which can be either a consonant or a vowel) or the syllable of the base is replaced by another phoneme or another syllable'* [Abbi et al., 1992: p. 20]. Examples from our data set of echo words include the following:

ધ્યાન રાખજો, **હાડકું બાડકું** તુટી ના જાય.

*Dhyan rakhjo, **hadku badku** tuti naa jaay.*

Be careful, not to break your bone.

वहाँ के रास्ते बहुत **तेढे मेढे** है ।

*Vaha ke raaste bahut **tedhe medhe** hai.*

The roads there are very uneven.

The aim of this study is to verify if expressives are actually being perceived differently from standard grammar or not. The perception of expressives was identifying by task functional Magnetic Resonance Imaging (*f*MRI). Difference in brain activation was obtained for response to sentences with and without expressives. The idea is to reduce the common aspects between expressives and non expressives as much as possible and focus only on the difference. Apart from that, reading and translation paradigms were used. It is our assumption that response to just reading a sentence should not be affected by the presence or absence of expressives. The act of reading is just vocalizing the symbols and should be almost independent of intricate elements of language. In translation, the meaning of the sentence must be understood and thus can be an indirect measure of perception. Two separate languages were used in the experimental paradigm to see if the perception difference of expressives was the same across the native and secondary language.

The next section details the methodology of the experiment with information on the experiment design and the analysis process. Results are presented in the third section followed by the discussion of the results in the fourth section. The fifth section discusses some of the limitations and challenges of the current study.

## 2. Methodology

*2.1 Functional experiment*

The functional experiment to identify the cognitive process of expressives involves two phases: a reading phase and a translation phase. Each phase consists of a mixed event related and block task *f*MRI experiment design. The entire functional experiment was divided into multiple blocks based on the language of the sentence and the presence or absence of expressives in the sentence. Each block can be categorized into one of the four categories – Gujarati expressives (GE), Gujarati non-expressives (GNE), Hindi expressives (HE) and Hindi non-expressives (HNE). For the reading phase, a sentence appeared on the screen in either Hindi or Gujarati and the participant had to read the sentence aloud. Instructions were given to the participants before entering the MRI scanner on how to speak from the diaphragm to minimize the head motion. (Head motion introduces motion artifacts in *f*MRI data.) Once the participant has finished reading a sentence, they would indicate with a button press and the next sentence would appear on the screen immediately after the button press. The button was pressed using the index finger of the right hand. A total of 5 sentences were presented together in a single block. At the end of each block, there was a blank screen with a small cross on the middle of the screen. The participants were instructed not to do anything during the blank screen and focus on the cross. The blank screen lasted for 10 seconds before the first sentence from the next block was displayed on the screen. Within each block, sentences from the same categories were displayed. The entire reading phase consists of 16 blocks, with 4 in each category. The order of the blocks was randomized for each participant. The graphic representation of the experiment paradigm is shown in Figure 1. The top portion shows one of the example screenshot for each group. The middle portion shows the timeline representation with different colors representing blocks of different categories. Each block consists of 5 separate sentences and the duration for each sentence is decided by the user with a button press.

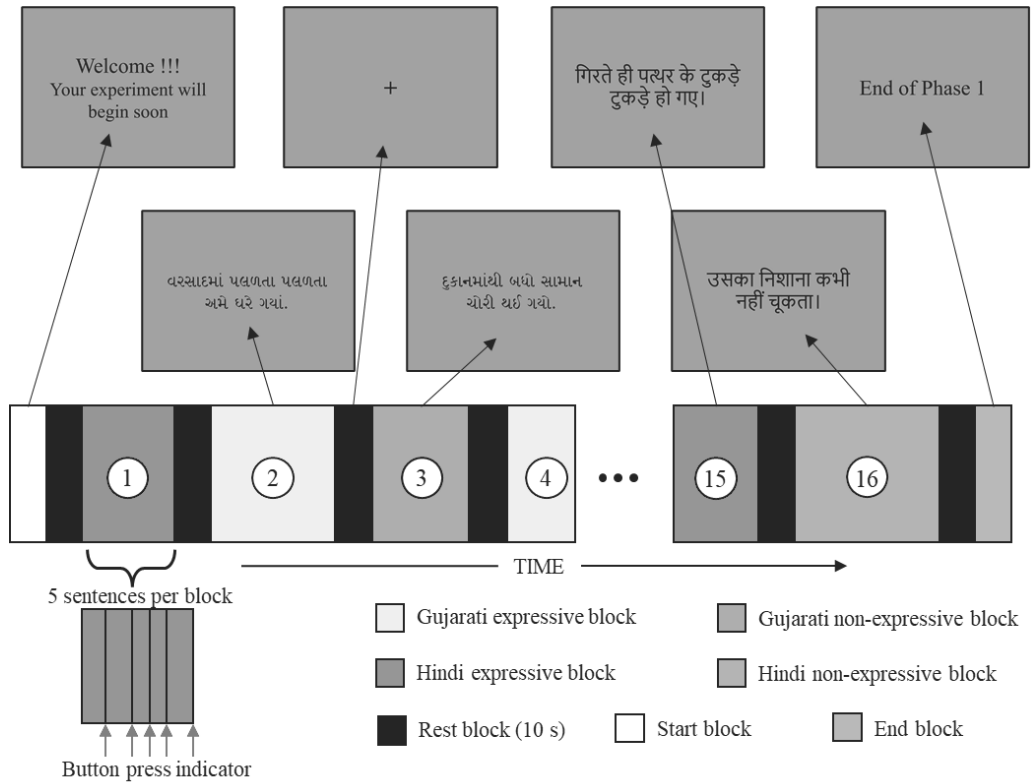

*Figure 1: Graphic representation for the experiment design.*

For the translation phase, the sentences were displayed on the screen in Hindi or Gujarati and the participants had to translate the sentences to English and speak out aloud the English translation. The participants were advised that there is no correct or incorrect translation and they had to translate the sentences the way they felt correct. Audio recording of the reading and translation phase was intended originally. However, it was decided prior to the scans that the loud noise of the MRI machine will not allow for good quality sound recording. Thus, no sound recording was made for the speech in both phases. A few of the participants opted only to participate in one language instead of both. For those participants, the experiment design was kept the same with only 2 categories (in the language of their choice) and 5 blocks per categories. Lastly, the duration for which each sentence appeared on the screen had an upper bound in case there was some error in button press or some participants deliberately didn't press the buttons. In case the button was not pressed, new sentence would appear on the screen after 10 seconds for reading phase and after 15 seconds for translation phase. The upper bound in time was kept assuming the slowest reading speed of 40 words per minute (w.p.m) (average reading speed being 120 – 150 w.p.m).

*2.2 Participants*

The participants were healthy adults who could read, understand and speak Hindi and/or Gujarati along with English. Prior to the experiment, approval was obtained from the Institution Review Board (IRB) of Texas Tech University and a signed consent was obtained from all the participants. All the participants were also screened for MRI safety through use of an MRI safety screening form. The total number of participants were 10 (4 F and 6 M) with an average (standard deviation) age of 31.2 (9.2) years. Apart from English, a total of eight participants knew both Gujarati and Hindi while one participant knew only Hindi, and one participant knew only Gujarati. One participant knew both Hindi and Gujarati but opted to participate only for Hindi experiment. Before the MRI scanning, the participants were asked to read from a fictional novel in Gujarati and Hindi for exactly one minute to measure their natural reading speed. The average reading speed across participants for Gujarati was 129 w.p.m (min = 100 w.p.m and max = 165 w.p.m) and for Hindi also was 129 w.p.m (min = 90 w.p.m and max = 175 w.p.m). No prior information was given to any participant about expressives to avoid any bias in the data. The participants also practiced on a demo task prior to entering the MRI scanner. All the sentences for the demo were without expressives, again to avoid any prior knowledge about the construct to be studied.

*2.3 MRI scan parameters*

The MRI scanning was performed on a 3T Siemens Skyra scanner. One high resolution anatomical scan and two separate functional scans were performed on each subject. The anatomical scans were performed using sagittal MPRAGE pulse sequence. The echo time (TE) and repetition time (TR) for anatomical scans is 2.49 millisecond and 1900 millisecond respectively. The resolution of sagittal images is 1 mm x 1 mm with a slice thickness of 0.9 mm. Two separate functional scans were acquired, one for the reading phase and one for the translation phase. The functional images were acquired using a multiband echo planar imaging (MB-EPI) pulse sequence [Setsompop et al., 2012]. Because of the multiband pulse sequence, six axial slices were acquired at a time allowing a lower TR value of 545 millisecond. For each functional volume, 48 axial slices of thickness 3.3 mm were acquired with an in-plane resolution of 3.25 mm x 3.25 mm. The experiment was self-paced thus different participants took different amount of time to complete it. The total number of functional volumes for reading phase range from 514 (4 min 40 sec) to 1036 (9 min 24 sec) and for translation phase range from 791 (7 min 11 sec) to 1490 (13 min 32 sec).

*2.4 Behavioral data analysis*

The first step in the analysis was to check if there is any difference in response time for expressives and non-expressives. The time to read/translate each sentence was known from the button press timing. Response time for each sentence was computed using the total time taken and the number of syllables in the sentence. The total number of syllables were calculated for all the sentences. To account for the differences in the syllable length, the read/translate times were weighted (divided) according to the syllable count. Hindi and Gujarati both are syllabary languages thus it is more suitable to weigh according to the number of syllables rather than number of words or characters. A paired t-test was conducted between the response times for reading and translating the sentences with and without expressives in it. First a paired t-test was conducted independently for all participants and then a group level paired t-test was conducted using the average response time per participant. For all the t-tests, a p value was computed and only the tests with p value less than to 0.05 were considered statistically significant.

Correlation was also computed between the average response time of the participants and their reading speed. Correlation was computed in all four categories for both reading and translation. Next, the difference between the reading and translation response times were analyzed. A paired t-test was performed between response timers of reading vs. translation for all participants. Independent t-tests were conducted for each of the 4 categories (GE, GNE, HE and HNE).

*2.5 Functional data analysis*

The functional data analysis begins with the preprocessing of the data. The preprocessing of the functional data was carried out using SPM12 toolbox [Ashburner et al., 2014] and in-house MATLAB script. The very first step was to convert and merge the raw DICOM files for each of the functional TR into a single Nifti file. The preprocessing steps that were then applied to the Nifti file were: motion correction, co-registration, spatial normalization, spatial segmentation, masking, temporal signal drift reduction, and spatial smoothing. The details about each of the preprocessing step and its need is given in Appendix A.

Preprocessing was followed by subject level analysis, also known as 1st level analysis. The first level analysis was performed using the GLM framework in the SPM12 toolbox using different regressors for each of the four categories. Motion parameters estimated in the preprocessing were also used as nuisance regressors. Subject wise activation regions were identified for different contrasts. These activation maps were combined across subjects to obtain the group level activation pattern. The group level inference was also performed using SPM 12 toolbox. The group level analysis for various contrasts were then analyzed individually to make inference about the perception of expressives and other aspects of the study. The spatial location of the activation regions were identified using the Brodmann atlas [Maldjian et al., 2003] and meta-analysis database from Neurosynth [https://neurosynth.org]. An in-depth description of the functional data analysis is given in Appendix B.

## 3. Results

*3.1 Behavioral data analysis*

The response times between expressives and non-expressives (E vs NE) were compared using a paired t-test at both subject and group level. There was no significant difference ($p < 0.05$) in the response times for participant level and group level analysis, except for some participants in Hindi reading. The detailed behavioral analysis results for response times are shown in Appendix C. For the reading phase, a strong correlation was observed between the response times and the reading speed across all participants. However, such correlation was not observed for translation phase. Finally, a significant difference ($p < 0.05$) in the response time was observed between reading and translation phase for all 4 categories. The detailed analysis for reading speed and response time is also given in Appendix C. In summary, the behavioral analysis did not yield any significant difference for response times between sentences with and without expressives for either language.

*3.2 Functional results*

The spatial clusters corresponding to the main contrast E vs NE are shown in Figure 2A. The contrast computes spatial regions showing significant activation for both reading and translation combined. The same contrast is computed for 3 separate cases: Gujarati alone, Hindi alone and both languages combined. The spatial brain activation regions corresponding to each of the three cases is shown in the figure with different color coding. Overlapping regions are represented by overlapping colors. The cluster level information for all the three cases is shown in Table 1. Table 1 also shows the cluster level information for the contrast NE vs E. The color coding for each of the case is as below:

The voxel with the most significant activation is within the occipital cortex. The clusters for both languages combined shows some similarities and some differences with the clusters of individual languages. For instance, the right inferior temporal gyrus shows significant activation for the contrast of Gujarati and both language combined and not for Hindi. Considering, most of the participants (9 out of 10) have Gujarati as their mother tongue (L1 language) it might be an indication of the processing region for L1 and not L2 (Hindi). Similar, occipital gyrus and DLPFC activation is observed only for Hindi and both languages combined. In differences, the caudate nucleus is only activated for Gujarati while parts of the supramarginal gyrus is only active for Hindi.

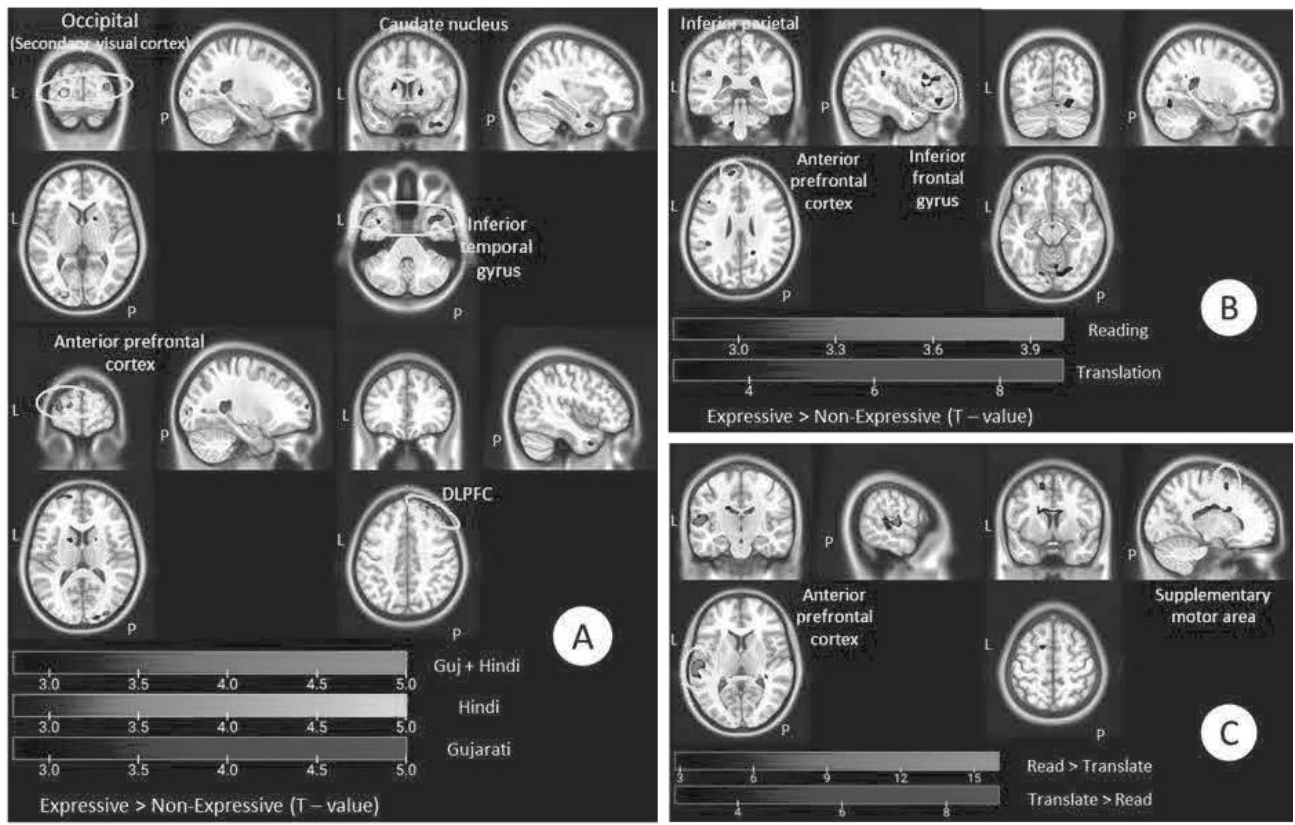

*Figure 2: (A) Spatial clusters for contrast E vs NE (reading + translation) for combination of different languages. (B) Spatial clusters of E vs NE contrast for reading and translation separately. (C) Spatial cluster for contrast R>T and T>R*

*Table 1: Cluster level information for all the contrasts shown in Figure 2. Different contrasts are indicated in **bold + italics**. Table also shows the activation region name, their Brodmann area, and the MNI coordinates (in mm) for the peak activation within the cluster along with its t-value.*

| Region Name | L/R | BA | MNI coordinate X | Y | Z | T | Region Name | L/R | BA | MNI coordinate X | Y | Z | T |
|---|---|---|---|---|---|---|---|---|---|---|---|---|---|
| ***Expressive > Non Expressives*** | *Gujarati & Hindi* | | *Read and Translate* | | | | ***Non Expressive > Expressive*** | *Gujarati & Hindi* | | *Read and Translate* | | | |
| Occipital (Secondary Visual) | R | 18 | 30 | -91 | 11 | 4.94 | Temporal gyrus | R | 21-22 | 42 | -28 | -4 | 9.08 |
| | | 18 | 18 | -97 | 14 | 3.66 | Middle temporal gyrus | R | 21 | 57 | -4 | -22 | 8.22 |
| DLPFC | R | 46 | 45 | 35 | 38 | 4.93 | Superior temporal gyrus | R | 22 | 42 | -58 | 20 | 6.86 |
| Anterior Prefrontal cortex | L | 10 | -30 | 62 | 14 | 4.42 | | | | | | | |
| | | 10 | -15 | 62 | 14 | 3.8 | | | | | | | |
| Occipital | L | 18 | -24 | -91 | 2 | 4.32 | | | | | | | |
| Inferior Temporal Gyrus | R | 20 | 33 | 9 | -37 | 3.83 | | | | | | | |
| ***Expressive > Non Expressives*** | *Gujarati* | | *Read and Translate* | | | | ***Expressive > Non Expressives*** | *Hindi* | | *Read and Translate* | | | |
| | R | 45 | 18 | 2 | -10 | 5.95 | Occipital | L | 18 | -24 | -91 | 2 | 6.06 |
| | | | 18 | 8 | 5 | 3.89 | Occipital | R | 18 | 21 | -94 | 11 | 3.4 |
| Inferior temporal gyrus | L | 20 | -39 | 5 | -34 | 5.04 | | | | 36 | -88 | 17 | 3.16 |
| | R | 36 | 30 | 5 | -37 | 4.67 | DLPFC | R | 46 | 42 | 35 | 41 | 5.51 |
| Caudate nucleus | L | | -15 | 5 | 14 | 4.5 | Supramarginal gyrus | L | 40 | -42 | -49 | 44 | 5.34 |
| ***Expressive > Non Expressives*** | *Gujarati + Hindi* | | *Reading* | | | | ***Read > Translate*** | *Gujarati + Hindi* | | | | | |
| Inferior Parietal | L | | -42 | -37 | 26 | 3.88 | Superior Temporal Gyrus | L | 22 | -60 | -19 | 8 | 32.37 |
| Anterior Prefrontal cortex | L | 10 | -12 | 62 | 23 | 3.6 | | R | 22 | 69 | -10 | 5 | 11.09 |
| ***Expressive > Non Expressives*** | *Gujarati + Hindi* | | *Translation* | | | | ***Translate > Read*** | *Gujarati + Hindi* | | | | | |
| | R | | 18 | -46 | 23 | 9.47 | | L | | -9 | -4 | 26 | 9.24 |
| Occipital | R | 18 | 24 | -70 | -13 | 7.38 | | R | | 12 | -7 | 26 | 7.49 |
| Occipital | L | 18 | -12 | -79 | -10 | 7.26 | DLPFC | L | 48 | -36 | 23 | 26 | 4.85 |
| Inferior frontal gyrus | L | 45 | -42 | 38 | 20 | 6.61 | | R | 48 | 30 | -34 | 14 | 4.24 |
| | L | | -42 | 41 | -16 | 5.58 | Sup motor | L | 6 | -15 | 5 | 59 | 4.2 |
| Inferior temporal gyrus | L | 20 | -42 | 5 | -28 | 5.5 | | | | | | | |

Apart from the language wise difference, the E vs NE contrast was also tested for reading and translation separately. Figure 2B shows the spatial activation regions while Table 1 shows the cluster level information. From the figure and the table, it can be observed that the difference between expressives for reading is very less compared to translation. There is also no overlapping region between reading and translation. To confirm the claim, two separate contrasts were tested – Reading > Translate and Translate > Reading. The spatial regions of both the contrasts are shown in Figure 2C and the cluster level information is shown in Table 1. Again, for reading, a smaller activation region is obtained as compared to translation. This suggests that the very act of translation involves more brain regions.

## 4. Discussion

The behavioral analysis suggests that there is no significant difference in the response times between expressives and non-expressives. For reading, it can be understood that sentences with expressives and non-expressives would be read in a similar fashion. However, no difference in the response times for translation was unexpected. One possible explanation for this could be the small compositional component of expressives in a sentence. In a typical sentence with 6-8 words, only a couple of words are expressives. Thus, the rest of the sentence would be translated as if there is no expressives present in them. Beyond this however, for some of the expressives, there is no corresponding translation in English which may induce a slightly larger change. Also, the response time for reading Gujarati and Hindi is similar while Hindi translation takes a bit longer than Gujarati translation. Given that 9 out of 10 participants were native Gujarati speakers, it is safe to assume that they would be relatively more proficient in Gujarati as compared to Hindi.

The main E vs NE contrast suggests the involvement of some of the brain regions in processing expressives differently from non-expressives. Usually the language related areas, Broca and Wernicke areas in paricular, are left lateralized [Zahn et al., 2000; Tie et al., 2014; Zhu et al., 2014]. However, many significant activation areas for the E vs. NE contrast are found in the right hemisphere. Right hemisphere activation is often seen when processing figurativeness and metaphorical aspect of the language [Diaz et al., 2011; Diaz & Hogstorm, 2011]. The right hemisphere activation of the E vs. NE contrast may suggest that expressives are being perceived as a more figurative and creative aspects of the language. Results also show a language dependent region of activation. In case of bilinguals, slightly different brain regions are involved in understanding L1 and L2 languages [Tie et al., 2014; Reverberi et al., 2018]. Most of the participants (9 out of 10) are native Gujarati speakers (L1) with Hindi and English as their secondary language (L2). Thus, it is very much possible that difference in the activation regions for Gujarati and Hindi is caused by differences in understanding L1 and L2.

In the case of L1, the E vs. NE contrast shows significant activation in the right inferior temporal gyrus and the caudate nucleus. Previous study shows the involvement of the caudate nucleus in the language switch circuitry [Wang et al., 2007; Hosoda et al., 2012], especially in forward switch from L1 to L2. Caudate nucleus is also shown to be involved in understanding the same concept in different languages [Abutalebi et al., 2008]. Activation in parts of the right inferior temporal gyrus and right temporal pole has been shown to be associated with creative and unique perception [Asari et al., 2018]. Asari et al., used vocal response from the participant which is similar to our experimental protocol as well. The same study also shows the involvement of the regions of the left anterior prefrontal cortex in creative mental activity. Activation of both the left inferior prefrontal cortex and the right inferior temporal gyrus for E vs NE contrast is yet another indication that expressives are being perceive as a more creative aspect of language.

For L2, the main regions of interest for E vs. NE contrast are within the occipital lobe, dorsolateral prefrontal cortex (DLPFC) and supramarginal gyrus. Some early fMRI studies have suggested the involvement of the occipital gyrus in word encoding [Kelly et al., 1998]. Recent studies have shown the involvement of the occipital gyrus areas in differentiating semantic concepts across different languages [Van de Putte et al., 2017] and language switch for bilinguals [Hosoda et al., 2012]. The DLPFC activation in the linguistic context is associated with verbal working memory [Veltman et al., 2003]. The act of translation requires the working memory aspect which also explains the DLPFC activation only for Translate > Reading contrast and not the other way around.

Comparison of the E vs. NE contrast for reading and translation also give an insight on how expressives are perceived by the brain. For reading, there isn't much difference in brain activity for expressives and non-expressives. The cluster size and T-values both are relative smaller compared to translation. Intuitively, reading a sentence with or without an expressive shouldn't be very different. However, for translation, the presence or absence of expressives do make a difference. Apart from the occipital and the inferior temporal gyrus, which have been discussed earlier, significant activation is observed in the inferior frontal gyrus (IFG). The IFG shows greater activation in encoding L2 language [Abutalebi et al., 2008.Reverberi et al., 2018]. IFG is also involved in language switch circuitry which is a crucial aspect of translation [Hosoda et al., 2012]. Apart from that, a study has shown higher activation of the IFG for semantically incorrect statements [Rüschemeyer et al., 2005]. Considering the comparative paucity of expressives in vernacular English as well as the degree of cultural knowledge associated with expressivity, it is possible that a direct word for word translation of expressives may result in a semantically incorrect statement resulting in higher activation of the IFG for expressive translation.

Activation in parts of the left IFG have been associated with creative thinking in various studies. Japardi et al., have shown a higher activation in the left IFG for Big-C group of people as compared to others [Japardi et al., 2018]. Big-C creative refers to individuals with exceptionally higher creative thinking abilities. Abraham et al., have shown the involvement of left IFG in creative conceptual expansion [Abraham et al., 2018] while Marron et al., have shown the involvement of IFG with free creative association [Marron et al., 2012]. Along with left IFG, both these studies also identify regions near the temporal poles to be involved in the creative tasks. Regions near the inferior temporal gyrus and temporal poles have also been associated with evaluation of aesthetic value for art [Kirk et al., 2009]. Thus, observing activations in the IFG, inferior temporal gyrus and some other regions discussed earlier may indicate that expressives are perceived as an aesthetic and creative aspect of grammar.

Another important contribution of this research is the fMRI for translation in bilinguals. Translation was compared against reading as a baseline. In the reading phase, the participants read out aloud the sentences displayed on the screen. In the translate phase, the participants read the sentence and then speak out the translation. Irrespective of whether expressives are present or not, the contrast of Translate > Read reveals the regions involved in the process of translation. The brain regions involved in the process of translation lie within the DLPFC and the supplementary motor area (SMA), as shown in Table 5. The function of DLPFC has been discussed earlier in verbal working memory. Along with DLPFC, SMA is also shown to be involved in verbal working memory [Veltman et al., 2003] and in forward language switch [Wang et al., 2007; Abutalebi et al., 2008]. On the other hand, the Read > Translate contrast shows activation in areas of superior temporal gyrus which is part of the famous Wernicke's area [Zhu et al., 2014]. The superior temporal gyrus is shown to be involved in detection of syntactic anomaly [Rüschemeyer et al., 2005]. This might suggest that less attention is focused on the syntactic correctness, or grammaticality, of a sentence when it is viewed for translation. Thus, the key neurological processes of translation can be summarized as follows: First, the given sentence is being read, but less attention is given to its exact correctness. Second the sentence is being stored in the verbal working memory. At the same time, the forward language switch regions mentally translate the sentence. Finally, the translated sentence is being spoken out. (Because both 'Read' and 'Translate' phase involves speaking, the vocal regions are not being identified in either Read > Translate or Translate > Read contrast.).

## 5. Limitations

The current study is an explorative study to find the neurological effects of expressives. As expressives are only a couple of words in a sentence, the difference between sentence with and without expressives is very small. This small difference leads to a smaller effect inside the brain when reading. For translation, the majority of difference lies in the translation of expressives when there is no corresponding translation for expressives in the other language. However, in translation as well the rest of the sentence can be translated easily. Thus, a difference of only a couple of words between sentences with and without expressives leads to relatively smaller difference in activation pattern. Another limitation is the inter subject variability. Variability factors like age, gender, handedness, reading speed, etc. can be controlled. It is a little difficult to control for abstract aspects of language proficiency like the knowledge about grammar and vocabulary. The inter subject variability along with smaller effect size makes it challenging to detect smaller activation regions across subjects with stringent statistical thresholds.

Limitation was also posed by the MRI hardware itself. First the participants were not allowed to move their head, especially lower jaw, while speaking. The participants could only speak with a lower amplitude than usual without getting uncomfortable. Second, the MRI scanner is very loud because of the constantly changing gradients. Altogether, the audio recording of the participants would be difficult to interpret. In such a scenario, the correctness of translation and reading cannot be quantified accurately. It can only be assumed that the participants performed the task correctly.

## 6. Conclusion

The results suggest that there are certain regions involved in processing the expressives differently from the non expressives. Also, most of the significant difference between expressives and non expressives is when translating the expressives into a language not having a direct correlation with expressives rather than just reading the expressives. Activation in some of the brain regions suggest that expressives are not processed completely at typical language processing regions. Also, the expressives may be perceived as a more creative and aesthetic part of language by brain. Apart from that, the cognitive process of language translation was also discussed.

*Appendix A: fMRI preprocessing*

Preprocessing is an important step for fMRI data analysis. Preprocessing reduces the effect of noise and artifact which may result in incorrect inference. During the scans, the participants tend to move a bit. The head motion introduces motion artifacts in the data. These motion artifacts are reduced by motion correction. The head motion of the participants is estimated and then functional volume at each TR are then corrected for the motion. Motion is corrected with 3D rigid body affine transform using 6 degrees of freedom (3 translational and 3 rotational). Motion correction of both phases of functional data is done together to avoid any spatial mismatch between the two phases. After motion correction, the high resolution anatomical and low resolution functional volumes are co-registered together. The co-registration step allows to map the results of the low-resolution functional images onto the high resolution anatomical volume after the analysis.

Every participant's brain is of different shape and size. To compare the results across multiple participants, all the brains should be of the same shape and size. The normalization step maps the functional volume of each participant onto a standard brain atlas. Thus, the brain shape and size for all participants is similar and functional results can be compared across multiple participants. The functional brain data for all participants was normalized to the MNI brain atlas [Fonov et al., 2009] with a spatial resolution of 3 mm x 3 mm x 3 mm. The normalized functional volume is then segmented into, primarily, gray matter, white matter, cerebrospinal fluid (CSF), skull and other soft tissue like skin and muscles. The gray matter, white matter and CSF segments are used to create a binary brain mask which eliminates all the area outside the brain from further analysis. Masking is important as it reduces the total number of voxels to be analyzed. A typical brain volume with a resolution of 3 mm x 3 mm x 3 mm has about 150,000 voxels. Out of these 150k voxels, only about 50,000 – 70,000 voxels are inside the brain region. Thus, discarding the extra voxels from further analysis increases computation and time efficiency.

After masking, only the voxels inside the brain region are used for further analysis. The next preprocessing step is the reduction of temporal signal drift. Due to scanner instabilities, there is a very low frequency temporal artifact introduced in the fMRI data. This artifact has a global effect and introduces noise in all the voxels across the brain. The temporal signal drift was estimated and reduced using a principal components analysis (PCA) technique [Parmar et al., 2019]. The last preprocessing step is spatial smoothing. In spatial smoothing a 3D Gaussian filter with a full width half maximum (FWHM) of 8 mm was used to smooth the functional data.

*Appendix B: Functional data analysis*

The preprocessed functional was analyzed to obtain spatial activation regions for different fMRI contrast. The analysis was done in two parts – subject level analysis and group level analysis. For subject level analysis the general linear model (GLM) framework was used to obtain information activation regions for each participant. Activation regions corresponding to multiple contrast were obtained for each subject. Then, in the group level analysis, the activation maps for all participants were combined.

The subject level analysis was performed using SPM12's *'First Level Analysis'*. Two separate GLM models were used for each participant – one for reading phase and one for translation phase. The timing for button press were used to generate onset time and duration for four categories (GE, GNE, HE and HNE). For each category three separate regressors were used in the design matrix of the GLM. The first regressor corresponds to the predicted response obtained by convolving the onset time and duration with canonical double gamma hemodynamic response function (HRF). The second and third regressors were obtained form time and dispersion derivative of the first regressor. The derivative regressors are used to account for any subject level variation in the HRF. The six estimated motion parameters were also used as nuisance regressors in the design matrix. The visualization of the entire design matrix for reading phase of participant 1 is shown in B1. In the visualization, the rows correspond to TRs while the columns correspond to regressors. The bright regions show higher amplitude while the dark regions show lower amplitude.

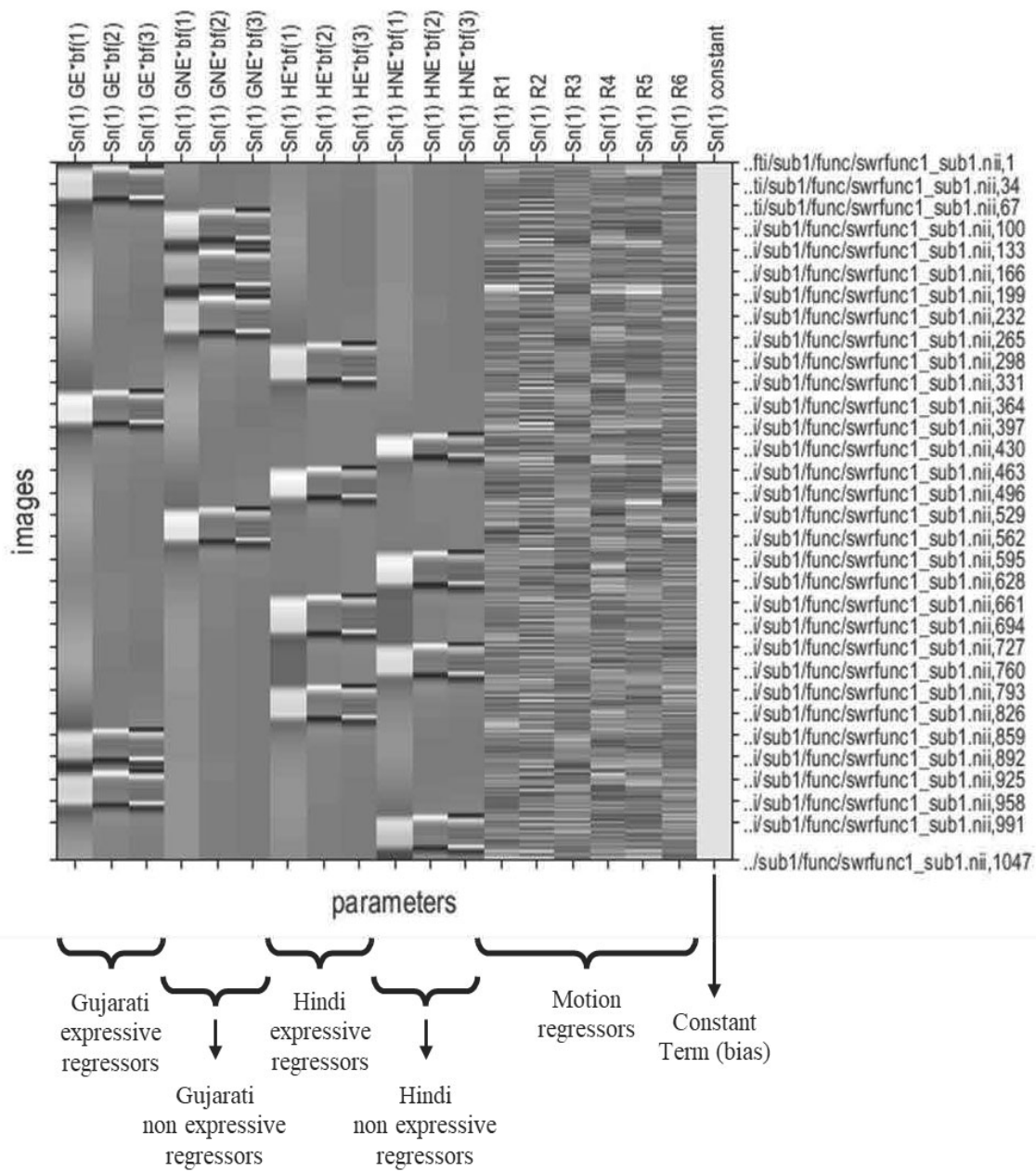

*Figure B1: Visualization of design matrix for first level analysis. The columns correspond to regressors while the rows correspond to time points. Higher amplitude is represented with brighter shade of gray while lower amplitudes are represented using darker shades of gray.*

$$Y = X\beta + \varepsilon$$

*Equation B1*

$$\beta \approx (X^T X)^{-1}(X^T Y)$$

*Equation B2*

The constructed design matrix is used with the GLM model to estimate the contribution (weights) of each of the regressors. The basic structure of the GLM model is shown in Equation B1. In the equation, **Y** is the matrix of observed time series. Each column is a time series from different voxel inside the brain. The matrix **X** is the design matrix constructed from the timing information and **β** is the matrix of weights that is to be estimated using Equation B2. The rows of **β** correspond to weights corresponding to a regressor while the columns correspond to voxels of the brain. After the estimation of **β**, the weights corresponding to each regressor are converted to 3D volume. In house MATLAB script was used to combine the 3D weight volumes for all subjects according to different contrasts. Some of the different contrasts tested are shown in Table B1. The volumes for different contrasts are saved for each participant.

The group level analysis was performed using SPM12's '*Second Level Analysis*' option. The input to the second level analysis were the saved contrast volumes for all of the participants. The second level analysis combines the contrast volume for all the participants and thresholds it according to the specified statistical significance. The spatial clusters were also thresholded according to the random field theory (RFT) to eliminate noisy single voxel clusters [Friston et al., 1994]. No significant group level clusters were identified at an uncorrelated p-value threshold of 0.001. Thus, the threshold was lowered to 0.005. The thresholded activation maps for all the contrast were saved for inference.

*Table B1: Weights for regressors corresponding to different contrasts.*

|  | READING | | | | TRANSLATION | | | |
| --- | --- | --- | --- | --- | --- | --- | --- | --- |
| **NAME** | GE | GNE | HE | HNE | GE | GNE | HE | HNE |
| *E > NE* | 1 | -1 | 1 | -1 | 1 | -1 | 1 | -1 |
| *NE > E* | -1 | 1 | -1 | 1 | -1 | 1 | -1 | 1 |
| *E > NE (Gujarati)* | 1 | -1 | 0 | 0 | 1 | -1 | 0 | 0 |
| *E > NE (Hindi)* | 0 | 0 | 1 | -1 | 0 | 0 | 1 | -1 |
| *E > NE (Read)* | 1 | 0 | 1 | 0 | -1 | 0 | -1 | 0 |
| *E > NE (Translate)* | -1 | 0 | -1 | 0 | 1 | 0 | 1 | 0 |
| *Read > Translate* | 1 | 1 | 1 | 1 | -1 | -1 | -1 | -1 |
| *Translate > Read* | -1 | -1 | -1 | -1 | 1 | 1 | 1 | 1 |

*Appendix C: Behavioral data analysis results*

Table C1 shows the calculated p-value for the subject level paired t-test. Each row corresponds to a subject while each column corresponds to a category. The four different categories of E vs NE are as follow – Gujarati reading, Hindi reading, Gujarati translation and Hindi translation. The category-subject pair for which there is a significant difference ($p < 0.05$) in the response time between expressive and non-expressives is **highlighted**. The participants that didn't performed the task of a particular category is highlighted in **red**. It can be observed that for most of the cases, there isn't a significant difference in response times between E vs NE. The above statement is confirmed from the group level paired t-test results. The group level p-value for all four categories are as follows: *p (Gujarati reading)* = 0.4142; *p (Hindi reading)* = 0.1908; *p*

*(Gujarati translation)* = 0.4207; and *p (Hindi translation)* = 0.5402. The p-values clearly suggest that there is no significant difference (*p<0.05*) in the response time between expressives and non expressives for either reading and translation. The response times across different subjects for both reading and translation is shown as boxplots in Figure C1.

*Table C1: Subject level p-value for paired t test*

| Participant | **Paired t-test for E vs NE** | | | |
| --- | --- | --- | --- | --- |
| | *Gujarati reading* | *Hindi reading* | *Gujarati translation* | *Hindi translation* |
| *1* | 0.239 | 0.120 | 0.485 | 0.595 |
| *2* | 0.064 | NA | 0.560 | NA |
| *3* | 0.607 | **0.001** | 0.806 | 0.162 |
| *4* | 0.622 | **0.001** | 0.935 | 0.898 |
| *5* | NA | 0.388 | NA | 0.070 |
| *6* | 0.259 | **0.019** | 0.940 | 0.257 |
| *7* | 0.327 | 0.644 | 0.635 | 0.281 |
| *8* | 0.495 | **0.026** | 0.907 | **0.040** |
| *9* | NA | 0.480 | NA | 0.874 |
| *10* | 0.172 | **0.047** | 0.703 | 0.180 |

The correlation is also computed between the response times and the reading speed of the participants. Figure C2 shows the plots for reading speed vs response time. For all 4 categories in reading, a strong negative correlation is observed between reading speed and response time. A negative correlation suggests that for higher reading speed participants will have a lower response time, which is obvious. However, the correlation between reading speed and response time for translation is different for Gujarati and Hindi. For Gujarati, there is a positive correlation suggesting that on average shorter sentences take a longer time to translate as compared to longer ones. This is true irrespective of the presence of expressives or not. On the other hand, there is no correlation in the reading speed and translation response time in Hindi.

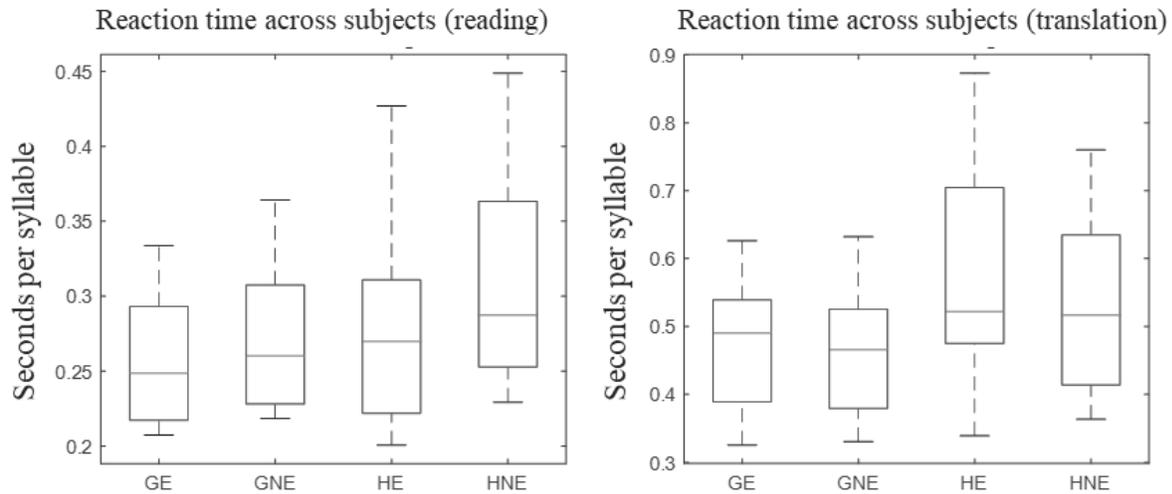

*Figure C1: Reaction time for reading and translation.*

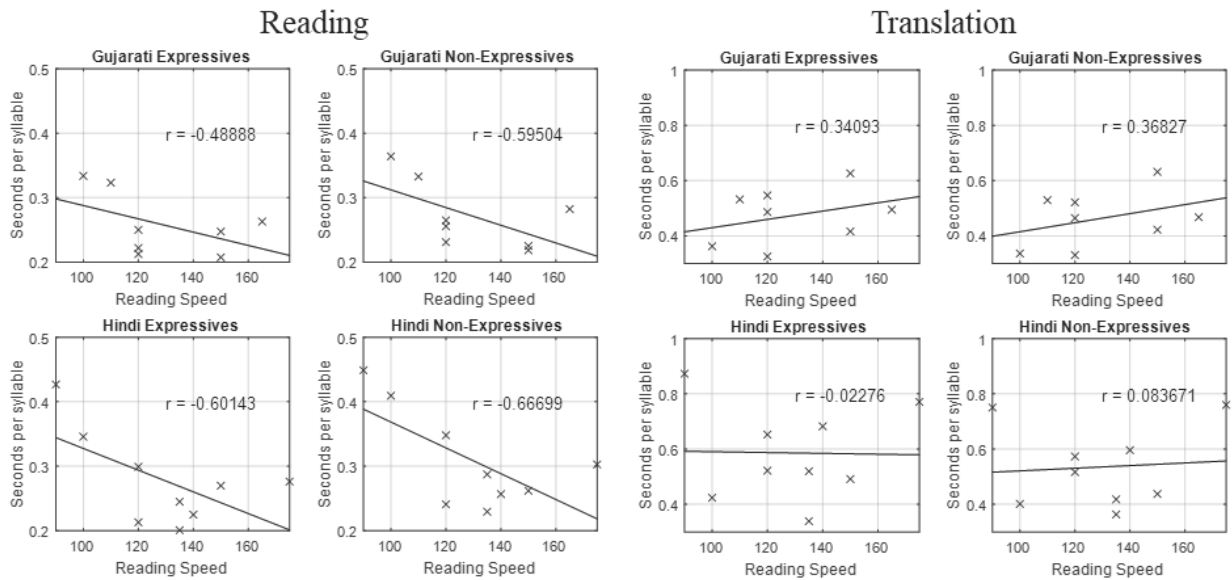

*Figure C2: Correlation between reading speed and response time*

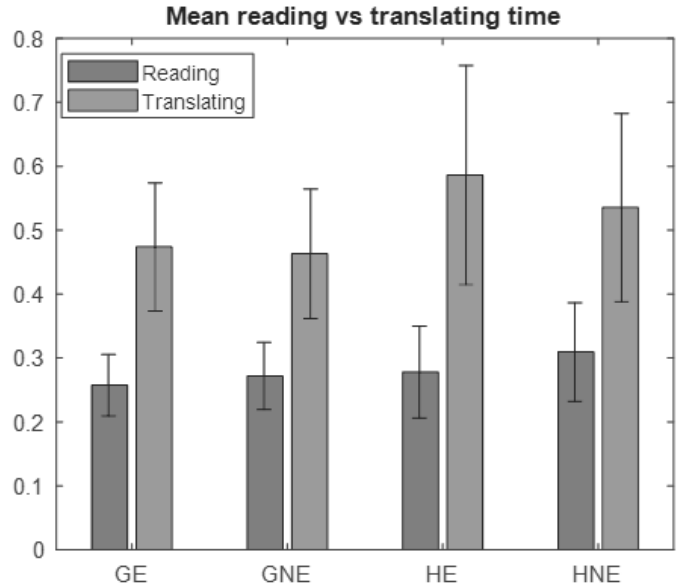

*Figure C3: Comparison of response times for reading and translation.*

The average response time for reading and translation for all 4 categories is shown in Figure C3. A paired t-test was also performed between the reading and translation response times. For all 4 cases, there is a significant increase ($p<0.05$) in the response times when going from reading to translation. The p-values for the paired t-test of all 4 categories are as follows: *p (Gujarati expressives)* = 0.0013; *p (Gujarati non-expressives)* = 0.0042; *p (Hindi expressives)* = 0.0004; and *p (Hindi non-expressives)* = 0.0016.

# References

Abbi, Anvita. "Contact, conflict and compromise: the genesis of reduplicated structures in South Asian languages." Dimock, Edward C., Kachru, Braj B., & Krishnamurti, Bh.(eds.) (1992): 131-148.

Abraham, Anna, Barbara Rutter, Trisha Bantin, and Christiane Hermann. "Creative conceptual expansion: A combined fMRI replication and extension study to examine individual differences in creativity." Neuropsychologia 118 (2018): 29-39.

Abutalebi, Jubin, Jean-Marie Annoni, Ivan Zimine, Alan J. Pegna, Mohamed L. Seghier, Hannelore Lee-Jahnke, François Lazeyras, Stefano F. Cappa, and Asaid Khateb. "Language control and lexical competition in bilinguals: an event-related fMRI study." Cerebral cortex 18, no. 7 (2008): 1496-1505.

Akita, Kimi. 2015. Sound symbolism. In Jan-Ola Östman & Jef Verschueren (eds). *Handbook of pragmatics*, 1–24. Amsterdam: John Benjamins.

_____ and Pardeshi (eds). 2019. *Ideophones, mimetics and expressives*. Amsterdam: John Benjamins.

Asari, Tomoki, Seiki Konishi, Koji Jimura, Junichi Chikazoe, Noriko Nakamura, and Yasushi Miyashita. "Right temporopolar activation associated with unique perception." NeuroImage 41, no. 1 (2008): 145-152.

Ashburner, John, Gareth Barnes, Chun-Chuan Chen, Jean Daunizeau, Guillaume Flandin, Karl Friston, Stefan Kiebel et al. "SPM12 manual." Wellcome Trust Centre for Neuroimaging, London, UK 2464 (2014).

Badenoch, Nathan and Osada, Toshiki. 2019. *A dictionary of Mundari expressives*. Tokyo: Research Institute for Languages and Cultures of Asia and Africa.

Benjamin, Christopher F., Patricia D. Walshaw, Kayleigh Hale, William D. Gaillard, Leslie C. Baxter, Madison M. Berl, Monika Polczynska et al. "Presurgical language fMRI: Mapping of six critical regions." Human brain mapping 38, no. 8 (2017): 4239-4255.

Berent, Iris, Hong Pan, Xu Zhao, Jane Epstein, Monica L. Bennett, Vibhas Deshpande, Ravi Teja Seethamraju, and Emily Stern. "Language universals engage Broca's area." PLoS One 9, no. 4 (2014): e95155.

Choksi, N. forthcoming. Expressives and the multimodal depiction of social types in Mundari. *Language in society,* 1-20. doi:10.1017/S0047404519000824

Diaz, Michele T., and Larson J. Hogstrom. "The influence of context on hemispheric recruitment during metaphor processing." Journal of cognitive neuroscience 23, no. 11 (2011): 3586-3597.

Diaz, Michele T., Kyle T. Barrett, and Larson J. Hogstrom. "The influence of sentence novelty and figurativeness on brain activity." Neuropsychologia 49, no. 3 (2011): 320-330.

Diffloth, Gérard. 1972. Notes on expressive meaning. *Chicago Linguistic Society* 8. 440–447.

_____. 1976. Expressives in Semai. *Oceanic Linguistics Special Publications* 13. 249–264.

_____. 1980. Expressive phonology and prosaic phonology in Mon-Khmer. In Theraphan L. Thongkum (ed.), *Studies in Mon-Khmer and Thai phonology and phonetics in honor of E. Henderson*, 49–59. Bangkok: Chulalongkorn University Press.

Dingemanse, Mark. 2012. Advances in the cross-linguistic study of ideophones. *Language and linguistics compass* 6(10). 654–672. DOI: https://doi.org/10.1002/lnc3.361.

_____. 2015. Ideophones and reduplication: Depiction, description, and the interpretation of repeated talk in discourse. *Studies in language* 39(4). 946–970. DOI: https://doi.org/10.1075/sl.39.4.05din.

_____. 2018. Redrawing the margins of language: Lessons from research on ideophones. *Glossa: a journal of general linguistics*, 3(1), p.4. DOI: http://doi.org/10.5334/gjgl.444

Friston, Karl J., Keith J. Worsley, Richard SJ Frackowiak, John C. Mazziotta, and Alan C. Evans. "Assessing the significance of focal activations using their spatial extent." Human brain mapping 1, no. 3 (1994): 210-220.

Fonov, Vladimir S., Alan C. Evans, Robert C. McKinstry, C. R. Almli, and D. L. Collins. "Unbiased nonlinear average age-appropriate brain templates from birth to adulthood." NeuroImage 47 (2009): S102.

Friederici, Angela D., Christian J. Fiebach, Matthias Schlesewsky, Ina D. Bornkessel, and D. Yves Von Cramon. "Processing linguistic complexity and grammaticality in the left frontal cortex." Cerebral cortex 16, no. 12 (2006): 1709-1717.

Haiman, John. 2019. *Ideophones and the evolution of language*. Cambridge University Press.

Hansen, Samuel J., Katie L. McMahon, and Greig I. de Zubicaray. "The neurobiology of taboo language processing: fMRI evidence during spoken word production." Social cognitive and affective neuroscience 14, no. 3 (2019): 271-279.

Hickok, Gregory. "The functional neuroanatomy of language." Physics of life reviews 6, no. 3 (2009): 121-143.

Hosoda, Chihiro, Takashi Hanakawa, Tadashi Nariai, Kikuo Ohno, and Manabu Honda. "Neural mechanisms of language switch." Journal of neurolinguistics 25, no. 1 (2012): 44-61.

Hubers, Ferdy, Tineke M. Snijders, and Helen De Hoop. "How the brain processes violations of the grammatical norm: An fMRI study." Brain and language 163 (2016): 22-31.

Ibarretxe-Antuñano, Iraide. 2017. Basque ideophones from a typological perspective. *Canadian Journal of Linguistics*, DOI: https://doi.org/10.1017/cnj.2017.8.

Iwasaki, Noriko Peter Sells & Kimi Akita (eds.). 2017. *The grammar of Japanese mimetics: Perspectives from structure, acquisition, and translation*. Routledge.

Jakobson, Roman. 1960. Closing statement: linguistics and poetics. In Thomas A. Sebeok, (ed.), *Style in language.* Cambridge, MA: MIT Press. Pp. 350-377.

Japardi, Kevin, Susan Bookheimer, Kendra Knudsen, Dara G. Ghahremani, and Robert M. Bilder. "Functional magnetic resonance imaging of divergent and convergent thinking in Big-C creativity." Neuropsychologia 118 (2018): 59-67.

Kelley, William M., Francis M. Miezin, Kathleen B. McDermott, Randy L. Buckner, Marcus E. Raichle, Neal J. Cohen, John M. Ollinger et al. "Hemispheric specialization in human dorsal frontal cortex and medial temporal lobe for verbal and nonverbal memory encoding." Neuron 20, no. 5 (1998): 927-936.

Ketteler, Daniel, Frank Kastrau, Rene Vohn, and Walter Huber. "The subcortical role of language processing. High level linguistic features such as ambiguity-resolution and the human brain; an fMRI study." NeuroImage 39, no. 4 (2008): 2002-2009.

Kirk, Ulrich, Martin Skov, Oliver Hulme, Mark S. Christensen, and Semir Zeki. "Modulation of aesthetic value by semantic context: An fMRI study." Neuroimage 44, no. 3 (2009): 1125-1132.

McCarthy, John J., Wendell Kimper, and Kevin Mullin. 2012. Reduplication in Harmonic Serialism. *Morphology* 22, 173-232.

Maldjian, Joseph A., Paul J. Laurienti, Robert A. Kraft, and Jonathan H. Burdette. "An automated method for neuroanatomic and cytoarchitectonic atlas-based interrogation of fMRI data sets." Neuroimage 19, no. 3 (2003): 1233-1239.

Marron, Tali R., Yulia Lerner, Ety Berant, Sivan Kinreich, Irit Shapira-Lichter, Talma Hendler, and Miriam Faust. "Chain free association, creativity, and the default mode network." Neuropsychologia 118 (2018): 40-58.

Nettekoven, Charlotte, Nicola Reck, Roland Goldbrunner, Christian Grefkes, and Carolin Weiß Lucas. "Short-and long-term reliability of language fMRI." NeuroImage 176 (2018): 215-225.ol

Reverberi, Carlo, Anna K. Kuhlen, Shima Seyed-Allaei, R. Stefan Greulich, Albert Costa, Jubin Abutalebi, and John-Dylan Haynes. "The neural basis of free language choice in bilingual speakers: Disentangling language choice and language execution." NeuroImage 177 (2018): 108-116.

Rüschemeyer, Shirley-Ann, Christian J. Fiebach, Vera Kempe, and Angela D. Friederici. "Processing lexical semantic and syntactic information in first and second language: fMRI evidence from German and Russian." Human brain mapping 25, no. 2 (2005): 266-286.

Sahin, Ned T., Steven Pinker, and Eric Halgren. "Abstract grammatical processing of nouns and verbs in Broca's area: evidence from fMRI." Cortex 42, no. 4 (2006): 540-562.

Scripture, Edward Wheeler. The elements of experimental phonetics. Vol. 22. C. Scribner's Sons, 1902.

Setsompop K, Cohen-Adad J, Gagoski BA, Raij T, Yendiki A, Keil B, Wedeen VJ, Wald LL. Improving diffusion MRI using simultaneous multi-slice echo planar imaging. NeuroImage. 2012 Oct 15;63(1):569-80.

Sulpizio, Simone, Michelle Toti, Nicola Del Maschio, Albert Costa, Davide Fedeli, Remo Job, and Jubin Abutalebi. "Are you really cursing? Neural processing of taboo words in native and foreign language." Brain and language 194 (2019): 84-92.

Tie, Yanmei, Laura Rigolo, Isaiah H. Norton, Raymond Y. Huang, Wentao Wu, Daniel Orringer, Srinivasan Mukundan Jr, and Alexandra J. Golby. "Defining language networks from resting-state fMRI for surgical planning—A feasibility study." Human brain mapping 35, no. 3 (2014): 1018-1030.

Tufvesson, Sylvia. "Analogy-making in the Semai sensory world." The senses and society 6: 86-95.

Van de Putte, Eowyn, Wouter De Baene, Marcel Brass, and Wouter Duyck. "Neural overlap of L1 and L2 semantic representations in speech: A decoding approach." NeuroImage 162 (2017): 106-116.

Veltman, Dick J., Serge ARB Rombouts, and Raymond J. Dolan. "Maintenance versus manipulation in verbal working memory revisited: an fMRI study." NeuroImage 18, no. 2 (2003): 247-256.

Voeltz, F. K. Erhard & Christa Kilian-Hatz (eds.),. (2001a). *Ideophones*. Amsterdam: John Benjamins. DOI: https://doi.org/10.1075/tsl.44.

Wang, Yapeng, Gui Xue, Chuansheng Chen, Feng Xue, and Qi Dong. "Neural bases of asymmetric language switching in second-language learners: An ER-fMRI study." NeuroImage 35, no. 2 (2007): 862-870.

Webster, Anthony K. 2008. 'To Give an Imagination to the Listener': The neglected poetics of Navajo ideophony. *Semiotica* 171. 343–365. DOI: https://doi.org/10.1515/SEMI.2008.081.

Williams, Jeffrey P., (ed.) The aesthetics of grammar: sound and meaning in the languages of Mainland Southeast Asia. Cambridge University Press. (2014).

_____. Expressive morphology in the languages of South Asia. Routledge. (2020).

_____. Expressives in the languages of Mainland Southeast Asia. Sidwell, Paul and Mathias, Jeremy, (eds.) (2021). Pages 811-824.

_____. Expressivity. Cambridge University Press. (forthcoming)

Zahn, Roland, Walter Huber, Eva Drews, Stephan Erberich, Timo Krings, Klaus Willmes, and Michael Schwarz. "Hemispheric lateralization at different levels of human auditory word processing: a functional magnetic resonance imaging study." Neuroscience letters 287, no. 3 (2000): 195-198.

Zhu, Linlin, Yang Fan, Qihong Zou, Jue Wang, Jia-Hong Gao, and Zhendong Niu. "Temporal reliability and lateralization of the resting-state language network." PLoS one 9, no. 1 (2014): e85880

Zwicky, Arnold and Geoffrey Pullum. "Plain morphology and expressive morphology." Proceedings of the Thirteenth Annual Meeting of the Berkeley Linguistics Society (1987): 330-340.